Twelfth International Multi-Conference on Information Processing-2016 (IMCIP-2016)

# A Harmony Search Based Wrapper Feature Selection Method for Holistic *Bangla* word Recognition

Supratim Das, Pawan Kumar Singh*, Showmik Bhowmik, Ram Sarkar and Mita Nasipuri

*Jadavpur University, West Bengal, Kolkata, India*

**Abstract**

A lot of search approaches have been explored for the selection of features in pattern classification domain in order to discover significant subset of the features which produces better accuracy. In this paper, we introduced a Harmony Search (HS) algorithm based feature selection method for feature dimensionality reduction in handwritten *Bangla* word recognition problem. This algorithm has been implemented to reduce the feature dimensionality of a technique described in one of our previous papers by S. Bhowmik *et al.*[1]. In the said paper, a set of 65 elliptical features were computed for handwritten *Bangla* word recognition purpose and a recognition accuracy of 81.37% was achieved using Multi Layer Perceptron (MLP) classifier. In the present work, a subset containing 48 features (approximately 75% of said feature vector) has been selected by HS based wrapper feature selection method which produces an accuracy rate of 90.29%. Reasonable outcomes also validates that the introduced algorithm utilizes optimal number of features while showing higher classification accuracies when compared to two standard evolutionary algorithms like Genetic Algorithm (GA), Particle Swarm Optimization (PSO) and statistical feature dimensionality reduction technique like Principal Component Analysis (PCA). This confirms the suitability of HS algorithm to the holistic handwritten word recognition problem.

## 1. Introduction

Feature dimensionality reduction can be defined as the method of diminishing the original feature set by detecting and eliminating the irrelevant ones while preserving adequate recognition rate if not less. The main objective of a good feature selection method is to lessen the feature computational cost while increasing the classifier efficiency. Feature selection has shown a considerable contribution in numerous applications such as pattern classification, multimedia information retrieval, data analysis, machine learning, medical data processing, and data mining[2]. The problem of dimensionality reduction, encompassing both extraction and selection of features, has been the subject of study in a diverse spectrum of fields. Improper estimations as well as high computational overhead are considered to be the major consequences of the enclosure of inappropriate and redundant features in the dataset. Improvement of interpretation of the formulated model, enhancement of learning procedure and reduction of the computational cost of feature dimension can be achieved by selecting pertinent subset of features. Due to the presence of significant amount of erroneous, unrelated, redundant or ambiguous features, selection of optimal features is usually necessary in almost all real world problems. Suppose a feature vector of size *n* is provided, the aim to select best features can be viewed as an exploration for an "optimal" subset of features out of $2^n$ candidate subsets[3]. The meaning of an optimal subset selection is a relative term with respect to the problem at hand. The schematic diagram of a general feature selection algorithm is shown in Fig. 1. Generally, the algorithms based on feature selection include two search strategies i.e., heuristic or random in order to avoid too much intricacy. However, the extent to select the optimal subset of features is often curtailed. At higher dimension, it is not easy to visualize the data; thus identifying features which may mislead the classifier during classification is not at all an easy task. This motivates most of the researchers to work in this domain.

*Corresponding author. Tel.: +91-9883033913.
*E-mail address:* pawansingh.ju@gmail.com

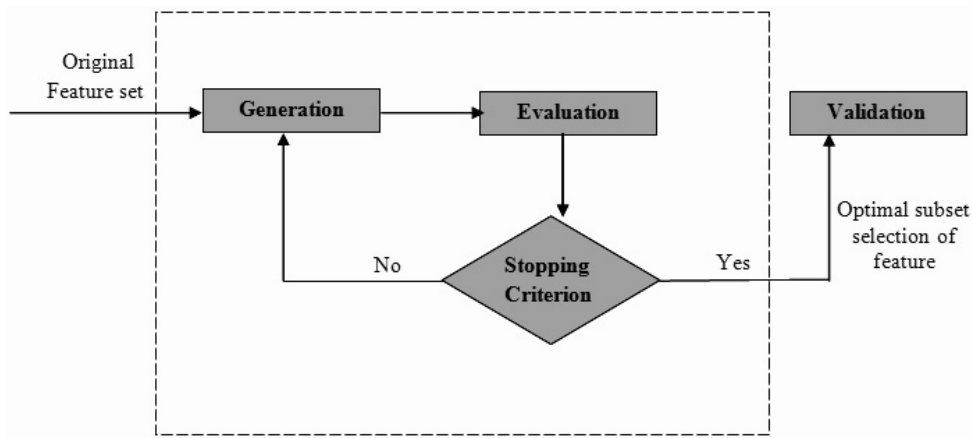

Fig. 1. Diagrammatic Representation of a General Feature Selection Method.

The literature survey on feature selection methodology confirms that a lot of researchers had opted for profusion of algorithms. These algorithms for feature selection can be generally pigeonholed into two approach *namely*, heuristic and meta-heuristic approaches. In general, these approaches can be further subdivided into two types of feature selection techniques: Filter method and Wrapper method[4]. In filter method, the features are ranked based on certain metrics, so, this method ignores the performance of classifier on the feature subset. Beside that it also overlooks the feature dependencies which may cause poor classification accuracy. On the other hand, the latter one is often used in conjunction with a machine learning algorithm, where the learning algorithm plays the part of the feature validation process. This method prefers to select an optimal feature subset which enhances the classification accuracy.

## 2. Related Work

In the literature, many heuristic algorithms are available for finding near–optimal solutions which are detailed in[5,6]. D. Nasien *et al.*[7] presented a comparative study on meta-heuristic algorithms *namely*, Genetic Algorithm (GA) and Ant Colony Optimization (ACO) to generate freeman chain code from handwritten character recognition to a digraph. The effectiveness of both the methods was evaluated on CEDAR (Center of Excellence for Document Analysis and Recognition) database. R. Azmi *et al.*[8] proposed a hybrid feature selection approach based on GA and simulated annealing algorithms. A database consisting of 100 samples for each 33 hand-printed *Farsi* characters was experimented using Bayesian classifier. A. Marcano-Cedeño *et al.*[9] presented a feature selection method based on Sequential Forward Selection (SFS) and Feed Forward Neural Network (FFNN) which assesses the prediction error as a selection criterion. D. Nasien *et al.*[10] proposed two heuristic algorithms that were based on randomized and enumeration algorithms to solve the problem. As a problem solving technique, the randomize algorithm made the random choices while the enumeration-based algorithm enumerated all possible candidates for a solution. The experiments were performed based on the chain code representation derived from established previous works of CEDAR dataset[7]. G. Abandah *et al.*[11] proposed a novel feature extraction approach for handwritten *Arabic* characters where the pre-segmented characters were first divided into its main body and secondary components. Next, the moment based features were individually extracted from the entire character as well as from the main body and the secondary components. Using multi-objective GA, the proficient subsets of feature were chosen. Based on their classification errors, these feature subsets were finally inspected using a Support Vector Machine (SVM) classifier. N. Das *et al.*[12] applied GA for selecting the optimal set of local regions. These regions then facilitated to extract an optimal set of features. The proposed methodology was evaluated on a dataset of handwritten *Bangla* digits. GA selected an optimal group of local regions which produced the best recognition performance with a SVM based classifier. A Roy *et al.*[13] implemented Artificial Bee Colony Optimization to identify the set of local regions which, in turn, provided the optimal set of discriminating features for the recognition of handwritten digits and characters. Initially, 8-directional gradient based feature values were estimated from each of the local regions produced by Center of Gravity (CG) based on Quad Tree partitioning approach. Subsequently, using this approach, at each level, the sampling process was carried out in every single region using SVM. C. De Stefano *et al.*[14] presented a novel GA-based feature selection algorithm in which feature subsets were evaluated with the help of an explicitly formulated separability index. This index measured the statistical properties of the feature subset and did not depend on any specific classification scheme. The proposed index represented an extension of the Fisher Linear Discriminant method and used covariance matrices for estimating how class probability distributions were quantized in the *N*-dimensional feature space. A key property of the approach was that any sort of a priori knowledge regarding the number of features was needed to select the feature subset. A. Roy *et al.*[15] proposed a new feature selection methodology based on the features combined class separability power. They firstly applied the framework of Axiomatic Fuzzy Set (AFS) theory which provided the rules for logic operations. These rules were required for the interpretation of the combinations of features from the fuzzy feature set. These combinational rules helped to determine the class separability power of the combined features and consequently, the most discriminative feature subset was chosen. F. G. Mohammadi *et al.*[16] introduced a new feature-based blind steganalysis method for the recognition of stego images from the cover images in JPEG images using a feature selection technique based on Artificial Bee Colony Optimization technique.

## 3. Motivation

After *Hindi, Bangla* is the most popular language in India[17]. *Bangla* is not only popular in *India* but it is also one of the most spoken language (7th position) in the world, which has nearly 230 million native speakers. This is the official and national language of Bangladesh. Apart from that, *Bangla* is also considered as an official language of some states in India *namely*, Tripura and West Bengal. *Bangla* script, with little modification is also used for writing languages such as *Manipuri, Assamese*, *Sylheti*. Despite of these facts, relatively less number of works[18,19] on word recognition system of handwritten *Bangla* script exists in the literature whereas other languages like *Arabic*, *French* and *English* etc. have many works on word recognition. Among the limited works on *Bangla* word recognition, most of the *Bangla* word recognition methods followed analytical approach[18]. But the analytical approach generally suffers from two major problems[20] first, "*segmentation ambiguity*" i.e. identifying the proper place in the word to segment it and the second one is "*variability of segment shape*" i.e. identification of each segment. These problems can be avoided with holistic approach[20] because it is completely a segmentation free approach. In holistic approach, feature values are extracted from the entire word image rather than the segment of the word image to denote it in the original feature space. Though holistic approach[21] has its own advantages over analytical approach, still it has achieved less attention from the researchers as it is always a huge challenge to design a suitable feature vector for the same. Thus, holistic *Bangla* word recognition is chosen as the ground for the present feature selection work.

Apart from this, it can be observed from the preceding literature survey that finding optimum number of features from a large feature space is difficult. For a large feature dimension, it is quite difficult to repetitively review the accuracy considering all possible feature subset. In fact, some common heuristic algorithms like hill climbing method can be applied by successively adding (or removing) feature to (from) the feature set. However, sometimes it gets stuck at local minima. Thus, the need to explore the search space using some heuristic search techniques is a major confront. Again, the limitation of using well-known statistical feature dimensionality method like PCA is that the principal component which is the largest eigen vector of the co-variance matrix generated is not often the optimal features in a lower dimension. The literature survey on *Bangla* word recognition problem also indicates that the exploration of optimum feature set for *Bangla* word recognition problem has got very little attention from research community in spite of its importance. However, heuristic search techniques, *namely*, Harmony Search (HS), Particle Swarm Optimization (PSO), and GA have been applied to reduce the complexity of character/digit recognition problem. In this paper, HS based search is described as this is still less explored although it become very popular in recent time due to its simplicity and stochastic behaviour (which helps us to escape from trapping in local optima). This is one of the most important reasons to assess the effectiveness of HS based wrapper feature selection method for realizing the improvement in the recognition accuracy of elliptical features based *Bangla* word recognition technique[1].

## 4. Proposed Work

Z. W. Geem *et al.*[22] pioneered a HS based meta-heuristic algorithm based on the concept which applies musical procedure to search a state of idealistic harmony. This is equivalent to find the optimality in an optimization process. Musicians play several notes using various distinct musical instruments and eventually discover the most appropriate combination of frequency for the finest tune. In the similar manner, the HS method selects the best combination among several existing solutions and finally optimizes the objective function. HS algorithm possesses numerous advantages[23] over other previously proposed meta-heuristic algorithms: (a) it desires less number of mathematical requirements and also it does not need to set initial values of the decision variables. (b) Due to the fact that the HS algorithm employs stochastic random searches, as a result, imitative information is also needless. (c) Considering all of the existing feature vectors, subsequently, HS algorithm produces a new feature vector. Hence, the compatibility of the HS algorithm is amplified by improvising these features leading to superior solutions.

The HS algorithm has five parameters: 1) the Harmony Memory Size (HMS); 2) Harmony Memory Considering Rate (HMCR); 3) Number of iterations; 4) Pitch Adjustment Rate (PAR); and 5) bandwidth. HMS corresponds to the number of solutions accumulated in the HM. HMCR is applied throughout the improvisation process in order to choose whether the variables of the solution should attain the value of any one present in HM. The value of HMCR lies in the range [0, 1] but in general 0.7 is taken as its experimental value. PAR is also utilized during the improvisation process to choose whether the variable of the solution should be altered to its neighbouring value. The value of PAR lies in the range [0.1, 0.5] but usually a value of 0.3 is preferred. The extent of modification is estimated by the bandwidth to change the solution from one neighbour to another. The linear adjustment of pitch is done in the following way:

$$x_{\text{new}} = x_{\text{old}} + P_{\text{band}} * \varepsilon \qquad (1)$$

where, $x_{\text{old}}$ is the existing pitch or solution from the HM, and $x_{\text{new}}$ is the pitch value after adjusting the existing pitch value. This basically generates a novel solution in the region of the prevailing solution by a small variation in the pitch value[22]. The constant, $\varepsilon$ is called a random number generator whose value lies in the range of $[-1, 1]$.

The pseudo code for HS algorithm is as follows:

```
Given Objective function f(x), x = (x₁, x₂, ......, x_d)^T where x = (x₁, x₂, ......, x_d)^T, we need to find the values x
which will maximize (or minimize) f(x)

Begin
Initialize PAR and bandwidth
Initialize HMCR, HM and Max number of iterations
Fill the HM with some probable solution vectors by choosing random harmonies between (min_val, max_val)
```

```
while ( t < Max number of iterations )
    //t indicating current iteration number must be initialized with 1
    {
        For each musician (x_i) do the following
        {
            Generate a random number m_1, where 0 ≤ m_1 ≤ 1
            if (m_1<HMCR)
            {
            Randomly choose a note from the existing domain of x_i
                Generate a random number m_2, where 0 ≤ m_2 ≤ 1
                if (m_2<PAR)
                {
                    Select the neighbouring value of already selected note, this range of choosing neighbouring value
                    depends on ε [here ε need to choose randomly].
                }
            }
            else
            {
                Randomly choose a value for x_i between (min_val, max_val)
            }
        } // end For

        Evaluate f(x) for newly generated solution vector.
        If newly generated solution vector is better than the worst solution vector in HM then replace the worst
        solution vector with newly generated solution vector.

        Increment t by 1
    } // end while
```

## 4.1 Example of feature selection using HS algorithm

All the data used in this example are hypothetical. Let us assume we have a total of 10 features i.e. $\{f_1, f_2, f_3, \ldots, f_{10}\}$ and we want to select the best 5 features among them. So, the total number of variables taken under consideration is 5. Here, our objective function is the classification accuracy evaluated by a given machine learning algorithm. The goal is to maximize this classification accuracy using HS algorithm. Let us consider the HMS to be 3.

|        | HMS                              | Classification Score |
|--------|----------------------------------|----------------------|
| Set #1 | $\langle f_1, f_5, f_3, f_2, f_9 \rangle$   | 40%                  |
| Set #2 | $\langle f_2, f_7, f_4, f_1, f_6 \rangle$   | 60%                  |
| Set #3 | $\langle f_{10}, f_1, f_2, f_3, f_7 \rangle$ | 70%                  |

During the generation of solution vector at initial stage, each musician must select distinct and random feature values. If we set HMCR=0.9 and PAR=0.1 then, $1^{st}$ musician will choose next note from the domain $\{f_1, f_2, f_{10}\}$ with probability 0.9 but still there exists a probability $(1 - 0.9)$ that $1^{st}$ musician may not choose from the range of all possible attributes values (i.e., $(f_1, f_2, f_3, \ldots, f_{10})$). Let us assume that $1^{st}$ musician has chosen $f_1$ value then, as per HS feature selection rule, $2^{nd}$ musician cannot choose $f_1$ value and so, he/she will have to choose the optimal feature from $\{f_5, f_7\}$ with probability 0.9. In this way, let us again suppose the $1^{st}$ musician has chosen $f_1$ and $2^{nd}$ musician has chosen $f_5$ and $3^{rd}$ musician has made a choice to select $f_4$. After making these choices, $3^{rd}$ musician can choose left ($f_2$) or right ($f_3$) value of $f_4$ as $3^{rd}$ musician has a total choices of 3 values $\{f_3, f_4, f_2\}$ and the probability of this selection to choose either left or right neighbor is 0.05 each. Therefore, at the end, $3^{rd}$ musician may select $f_2$ as the final value.

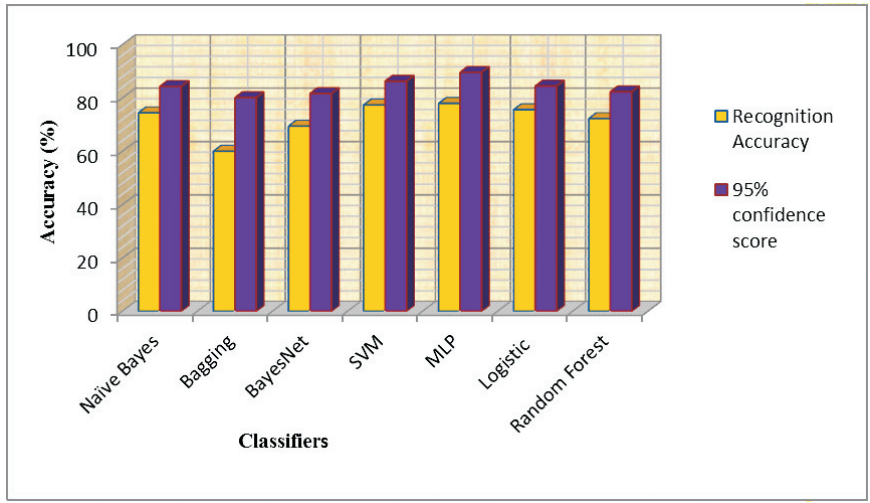

Fig. 2.  Illustration of Recognition Accuracies and its Corresponding 95% Confidence Scores using Seven Classifiers on Handwritten *Bangla* Word Recognition Database.

## 5. Experimental Study and Discussion

The work described in[1] has been evaluated on a dataset consisting of 1020 handwritten words. This database contains 20 distinct word classes where each class contributes an exactly 51 number of handwritten words. A total of 680 word images have been used to train the scheme whereas the remaining 340 images has been taken for testing the same. A set of 65 elliptical features has been extracted from each of the word images and tested using seven classifiers *namely*, Naïve Bayes, Bagging, BayesNet, SVM, MLP, Logistic and Random Forest. The recognition accuracies and their corresponding confidence scores are shown in Fig. 2. It can be witnessed that the highest recognition accuracy of 81.37% is attained using MLP classifier.

It is difficult to decide the optimum number of features prior to the feature selection. So, the HS based feature selection technique is tested by selecting 15%, 30%, 45%, 60%, 75% and 90% of the original feature vector. These percentages are rounded off to the lowest nearest integer. Here, the objective function is the obviously classification accuracy given by MLP classifier. As S. Bhowmik *et al.*[1] had achieved the classification accuracy of 81.37% using MLP classifier, so we have chosen MLP classifier for the current experimentation. To develop a trained network of MLP classifier, 1000 runs of back propagation learning algorithm with learning rate ($\eta$) = 0.3, momentum term ($\alpha$) = 0.4 and adjustment factor = 0.7 are executed for different number of neurons in its hidden layer. The corresponding classification accuracies achieved against the percentage of elliptical features taken for experimentation using 3-fold cross validation scheme are shown in Fig.3. For the present work, the optimal values of HMCR and PAR have been experimentally set to 0.7 and 0.3. Keeping these values fixed, a comparison between the different values of HMS and varied number of iterations has been performed which is detailed in Table 1. It can be observed from Table 1 that for HMS=20 and 10 number of iterations, the algorithm attains the highest recognition accuracy of 90.29% which is much higher than base work[1] considered here.

For the present work, the optimization performance of HS algorithm is also compared with two popular meta-heuristic algorithms such as GA[24], and PSO[25]. The GA is implemented to contain 20 chromosomes and made to iterate for 100 generations in each trial. The crossover rate and mutation rate are set to 1.0 and 0.1 respectively. For PSO algorithm, the 20 particles is chosen for the experimental purpose. The two factors $rand_1$ and $rand_2$ are random numbers lying between (0, 1) whereas $C_1$ and $C_2$ are acceleration (learning) factors, with $C_1 = C_2 = 2$. The inertia weight $w$ is taken as 0.9 and the maximum number of iterations used in our PSO is 100. The comparison is measured with respect to three parameters: (a) optimal feature subset, (b) classification accuracy, and (c) execution time. Optimal feature subset denotes the number of features required to achieve the optimal maximum recognition

Table 1. Recognition Accuracies of the Present HS Algorithm for Different Parameter Values of HMS and Number of Iterations (best case is shaded in Gray and Styled in bold).

| Number of Iterations | Recognition accuracy (%) HMS | | | | |
|---|---|---|---|---|---|
| | 10 | 20 | 30 | 40 | 50 |
| 10 | 89.31 | **90.29** | 90.23 | 90.18 | 90.13 |
| 20 | 89.37 | 90.11 | 90.18 | 89.66 | 90.02 |
| 30 | 88.89 | 89.21 | 90.11 | 90.06 | 89.58 |
| 40 | 89.55 | 90.18 | 89.45 | 89.22 | 89.96 |
| 50 | 88.48 | 90.11 | 90.02 | 90.18 | 89.42 |

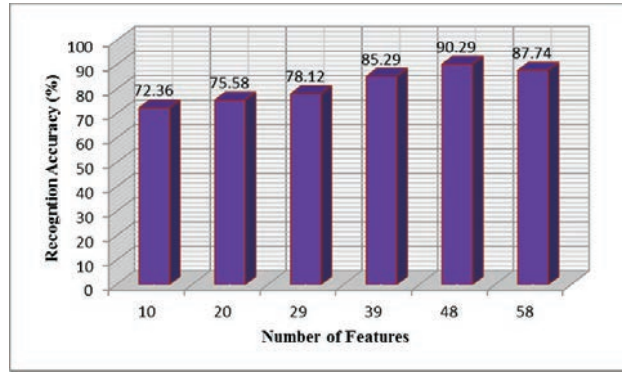

Fig. 3. Comparison of Recognition Accuracies of the Proposed Technique for the Number of Optimal Features.

Table 2. Comparison of Best Case of the Optimization Performance of HS Algorithm with GA and PSO Algorithms.

| Optimization Algorithm | Optimum Feature Subset | Classification Accuracy (%) | Execution Time (secs.) |
|---|---|---|---|
| GA | 45 | 84.65 | 1509.25 |
| PSO | 40 | 85.19 | 1248.89 |
| **HS** | **48** | **90.29** | **944.75** |

accuracy. The classification accuracy of the present work is evaluated as follows:

$$\text{Classification Accuracy}(\%) = \frac{\text{Number of correctly classified words}}{\text{Total number of words}} \times 100(\%) \qquad (2)$$

The execution time is measured as the time required to run the given optimization algorithm until the classification accuracy is realized using MLP classifier with 3-fold cross validation scheme. The comparison enlisted in Table 2 confirms that the present HS algorithm not only achieves the highest classification accuracy but also involves the minimum computational time. Despite the fact that PSO selects comparatively less number of features than the HS algorithm but the latter achieves about 5% more recognition rate than the former while taking less computational time. Again, a well-known statistical feature dimensionality technique named PCA[26] is also implemented and evaluated on the same dataset and it is found that PCA utilizes a feature dimensionality of 40 yielding a maximum recognition accuracy of 84.97%. Therefore, it is quite clear from the preceding observations that the HS algorithm outperforms some pioneered optimization algorithms used mainly for feature selection.

## 6. Conclusions

This paper has examined the HS algorithm for the handwritten holistic word recognition of *Bangla* script with minimum number of features. The proposed approach evolved upon the HM which stores adaptively updated solutions during the evolution. The approach has been tested on a limited dataset of 1020 *Bangla* handwritten words benchmark in the literature. For obtaining the optimal classification accuracy, a series of experiments have been performed to assess the effectiveness of HS algorithm with different parameter settings and the algorithm achieves an impressive classification accuracy of 90.29% which is almost about 9% more than the earlier attained result. The improved result justifies the need of feature selection for this type of pattern classification problem where both local and global features are computed without knowing the exact importance of the features.

The future scope of the current work will be to automatically tune and select the parameters of the algorithm according to the given problem. Setting of an ideal stopping criterion for the HS algorithm would also be beneficial. Additionally, a scheme may be formulated in order to dynamically identify the size of the HM, which will influence both convergence rate as well as the search parameters. The future goal will also aim to incorporate a comparative study of this result with other state-of-the-art meta-heuristic methods proposed in the literature.

## References

[1] S. Bhowmik, S. Malakar, R. Sarkar and M. Nasipuri, Handwritten *Bangla* Word Recognition using Elliptical Features, In *International Conference on Computational Intelligence and Communication Networks (CICN)*, pp. 257–261, (2014).
[2] I. Guyon and A. Elisseeff, Special Issue on Variable and Feature Selection, In *Journal of Machine Learning Research*, vol. 3, pp. 1157–1182, (2008).
[3] M. L. Raymer, W. F. Punch, E. D. Goodman, L. A. Kuhn and A. K. Jain, Dimensionality Reduction using Genetic Algorithms, In *IEEE Transactions on Evolutionary Computation*, vol. 4, no. 2. pp. 164–171, (2000).
[4] T. Fagbola, S. Olabiyisi and A. Adigun, Hybrid GA-SVM for Efficient Feature Selection in E-mail Classification, In *Computer Engineering and Intelligent Systems*, vol. 3, no. 3, pp. 17–28, (2012).
[5] N. Kwak and C. H. Choi, Input Feature Selection for Classification Problems, In *IEEE Transactions on Neural Networks*, vol. 13, no. 1, pp. 143–159, (2002).
[6] P. Somol, P. Pudil and J. Kittler, Fast Branch and Bound Algorithms for Optimal Feature Selection, In *IEEE Transactions on Pattern Analysis and Machine Intelligence*, vol. 26, no. 7, pp. 900–912, (2004).
[7] D. Nasien, H. Haron and S. S. Yuhaniz, Metaheuristics Methods (GA & ACO) for Minimizing the Length of Freeman Chain Code from Handwritten Isolated Characters, In *World Academy of Science Engineering and Technology*, vol. 62, article 41, pp. 230–235, February (2010).
[8] R. Azmi, B. Pishgoo, N. Norozi, M. Koohzadi and F. Baesi, A Hybrid GA and SA Algorithms for Feature Selection in Recognition of Hand-printed Farsi Characters, In *International Conference on Intelligent Computing and Intelligent Systems (ICIS)*, vol. 3, pp. 384–387, (2010).
[9] A. Marcano-Cedeño, J. Quintanilla-Domínguez, M. G. Cortina-Janchs and D. Andina, Feature Selection using Sequential Forward Selection and Classification Applying Artificial Metaplasticity Neural Network, In $36^{th}$ *Annual Conference on IEEE Industrial Electronics Society (IECON-2010)*, pp. 2845–2850, (2010).
[10] D. Nasien, H. Haron and S. S. Yuhaniz, The Heuristic Extraction Algorithm for Freeman Chain Code of Handwritten Character, In *International Journal of Experimental Algorithms (IJEA), Publisher: CSC Press, Computer Science Journals*, vol. 1, issue 1, ISSN: 2180-1282, pp. 1–20, (2011).
[11] G. Abandah and N. Anssari, Novel Moment Features Extraction for Recognizing Handwritten Arabic Letters, In *Journal of Computer Science*, vol. 5(3), pp. 226–232, (2009).
[12] N. Das, R. Sarkar, S. Basu, M. Kundu, M. Nasipuri and D. K. Basu, A Genetic Algorithm Based Region Sampling for Selection of Local Features in Handwritten Digit Recognition Application, In *Applied Soft Computing*, vol. 12, pp. 1592–1606, (2012).
[13] A. Roy, N. Das, R. Sarkar, S. Basu, M. Kundu and M. Nasipuri, Region Selection in Handwritten Character Recognition using Artificial Bee Colony Optimization, In *Proceedings of $3^{rd}$ International Conference on Emerging Applications of Information Technology (EAIT)*, pp. 183–186, (2012).
[14] C. De Stefano, F. Fontanella, C. Marrocco and A. Scotto di Freca, AGA-based Feature Selection Approach with An Application to Handwritten Character Recognition, In *Pattern Recognition Letters*, vol. 35, pp. 130–141, (2014).
[15] A. Roy, N. Das, R. Sarkar, S. Basu, M. Kundu and M. Nasipuri, An Axiomatic Fuzzy Set Theory Based Feature Selection Methodology for Handwritten Numeral Recognition, In *Proceedings of $48^{th}$ Annual Convention of Computer Society of India*, vol. I, AISC248, pp. 133–140, (2014).
[16] F. G. Mohammadi and M. S. Abadeh, Imagesteg Analysis using a Bee Colony Based Feature Selection Algorithm, In *Engineering Applications of Artificial Intelligence*, vol. 31, pp. 35–43, (2014).
[17] http://www.languageinindia.com/feb2011/vanishreemastersfinal.pdf Retrieved 2016-03-05.
[18] N. Das, S. Pramanik, S. Basu, P. K. Saha, R. Sarkar, M. Kundu and M. Nasipuri, Recognition of Handwritten Bangla Basic Characters and Digits using Convex Hull Based Feature Set, In *Proceedings of International Conference on Artificial Intelligence and Pattern Recognition (AIPR)*, pp. 380–386, (2009).
[19] N. Das, K. Acharya, R. Sarkar, S. Basu, M. Kundu and M. Nasipuri, A Benchmark Image Database of Isolated Bangla Handwritten Compound Characters, In *International Journal on Document Analysis and Recognition*, vol. 17, no. 4, pp. 413–431, (2014).
[20] S. Bhowmik, S. Polley, Md. G. Roushan, S. Malakar, R. Sarkar and M. Nasipuri, A Holistic Word Recognition Technique for Handwritten Bangla Words, In *International Journal of Applied Pattern Recognition*, vol. 2, no. 2, pp. 142–159, (2015).

[21] A. Acharyya, S. Rakshit, R. Sarkar, S. Basu and M. Nasipuri, Handwritten Word Recognition using MLP based Classifier: A Holistic Approach, In *International Journal of Computer Science*, issues 10(2), pp. 422–427, (2013).
[22] Z. W. Geem, J. H. Kim and G. V. Loganathan, A New Heuristic Optimization Algorithm: Harmony Search, In *Simulation*, vol. 76, no. 2, pp. 60–68, (2001).
[23] K. S. Lee and Z. W. Geem, A New Structural Optimization Method Based on the Harmony Search Algorithm, *Journal of Computers and Structures*, vol. 82, pp. 781–798, (2004).
[24] J. H. Holland, Adaptation in Natural and Artificial Systems, In *University of Michigan Press*, New York, (1975).
[25] J. Kennedy and R. C. Eberhart, Particle Swarm Optimization, In *Proceedings of IEEE International Conference on Neural Networks*, pp. 1942–1948, (1995).
[26] K. Pearson, On Lines and Planes of Closest Fit to Systems of Points in Space, In *Philosophical Magazine*, vol. 2(11), pp. 559–572, (1901).